\documentclass{article} 
\usepackage{iclr2025_conference,times}

\usepackage{amsmath,amsfonts,bm}

\def\eqref#1{equation~\ref{#1}}

\def\1{\bm{1}}

\DeclareMathAlphabet{\mathsfit}{\encodingdefault}{\sfdefault}{m}{sl}
\SetMathAlphabet{\mathsfit}{bold}{\encodingdefault}{\sfdefault}{bx}{n}

\usepackage{hyperref}
\usepackage{url}
\usepackage{microtype}
\usepackage{graphicx,wrapfig,lipsum}
\usepackage{subfigure}
\usepackage{booktabs} 

\iclrfinaltrue

\newif\ifshowcomments
\showcommentstrue 
\ifshowcomments
\newcommand {\konstantin}[1]{{\color{red}\sf{[Konstantin: #1]}}}
\newcommand {\ivan}[1]{{\color{purple}\sf{[Ivan: #1]}}}
\definecolor{gold}{HTML}{DAA520}
\newcommand {\leo}[1]{{\color{gold}\sf{[Léo: #1]}}}
\definecolor{darkgreen}{HTML}{20b020}
\newcommand {\damian}[1]{{\color{darkgreen}\sf{[Damian: #1]}}}
\newcommand {\goeran}[1]{{\color{blue}\sf{[Göran: #1]}}}
\else
\newcommand {\konstantin}[1]{}
\newcommand {\ivan}[1]{}
\newcommand{\leo}[1]{}
\newcommand{\damian}[1]{}
\newcommand{\goeran}[1]{}
\fi

\title{Structure Is Not Enough: Leveraging Behavior for Neural Network Weight Reconstruction}

\author{Léo Meynent$^1$, Ivan Melev$^2$, Konstantin Schürholt$^1$, Göran Kauermann$^2$, Damian Borth$^1$\\
$^1$AIML Lab, University of St. Gallen, Switerland\\
$^2$Department of Statistics, Ludwig-Maximilians-University Munich, Germany\\
Correspondence: \texttt{leo.meynent@unisg.ch}
}

\begin{document}

\maketitle

\begin{abstract}
The weights of neural networks (NNs) have recently gained prominence as a new data modality in machine learning, with applications ranging from accuracy and hyperparameter prediction to representation learning or weight generation. One approach to leverage NN weights involves training autoencoders (AEs), using contrastive and reconstruction losses. This allows such models to be applied to a wide variety of downstream tasks, and they demonstrate strong predictive performance and low reconstruction error. However, despite the low reconstruction error, these AEs reconstruct NN models with deteriorated performance compared to the original ones, limiting their usability with regard to model weight generation. 
In this paper, we identify a limitation of weight-space AEs, specifically highlighting that a \textit{structural} loss, that uses the Euclidean distance between original and reconstructed weights, fails to capture some features critical for reconstructing high-performing models. We analyze the addition of a \textit{behavioral} loss for training AEs in weight space, where we compare the output of the reconstructed model with that of the original one, given some common input. We show a strong synergy between structural and behavioral signals, leading to increased performance in all downstream tasks evaluated, in particular NN weights reconstruction and generation.\looseness-1
\end{abstract}

\section{Introduction}

\begin{wrapfigure}{r}{0.50\textwidth}
\begin{center}
\vskip -0.6in
\centerline{\includegraphics[width=0.45\textwidth]{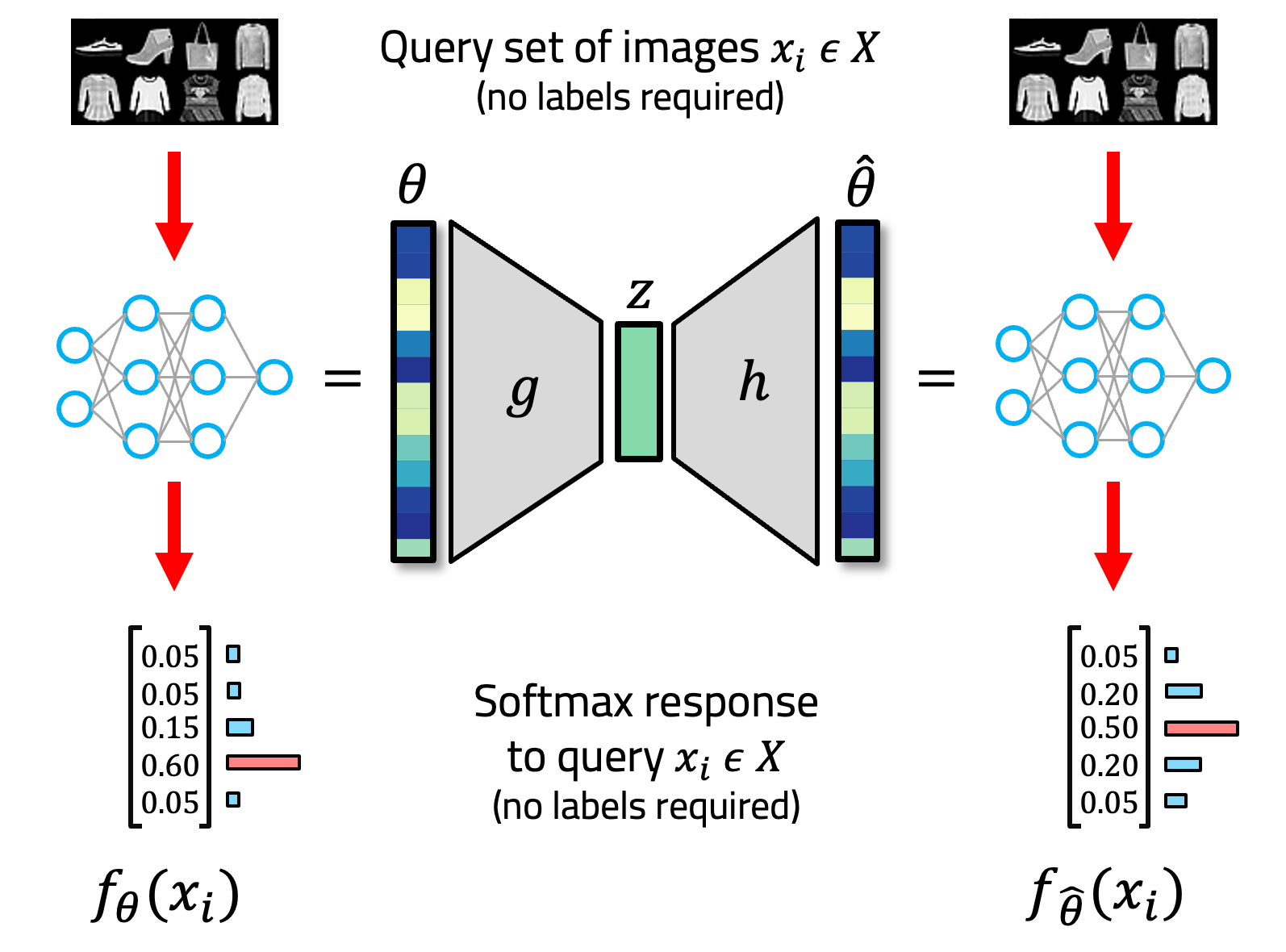}}
\vskip -0.15in
\caption{Illustration of our behavioral loss. Previous works have used a \textit{structural} loss, comparing NN weights with their reconstruction using the Euclidean distance. In that setting however, they failed to reconstruct models that perform as a well as the original. We propose combining it with a \textit{behavioral} loss, where we use \textit{queries} to compare the outputs of the NN and its reconstruction. Since this loss does not attempt to make the reconstructed model accurate with regard to a ground truth, rather to match its behavior with that of the original, the queries do not need to be labeled.\looseness-1}
\vskip -0.45in
\label{fig:overview}
\end{center}
\end{wrapfigure}

The weights of trained neural network (NN) models have recently become themselves a domain for research and machine learning, named \textit{weight space learning}~\citep{kofinasGraphNeuralNetworks2023, limGraphMetanetworksProcessing2023, navonEquivariantArchitecturesLearning2023, schurholtScalableVersatileWeight2024, zhouPermutationEquivariantNeural2023}. 
Since the weights are structured by the training process, they contain rich information on the data, their generating factors, and also model performance \citep{unterthinerPredictingNeuralNetwork2020,martinPredictingTrendsQuality2021}.
This opens up the opportunity to analyze NNs by investigating their weights. Understanding the inherent structure in trained weights might further lead to better initializations \citep{narkhede2022review}, model merging \citep{chou2018unifying}, or identification of lottery tickets~\citep{frankle2018lottery}.

Several methods have been proposed to learn from NN weights for a variety of downstream tasks. Weight derived features have been used for \textit{discriminative} downstream tasks, such as predicting model performance of training hyperparameters~\citep{unterthinerPredictingNeuralNetwork2020,eilertsenClassifyingClassifierDissecting2020,martinPredictingTrendsQuality2021,schurholtSelfSupervisedRepresentationLearning2021}.
In reverse, others have developed methods for \textit{generative} downstream tasks, i.e., the generation of synthetic NN weights. For example, HyperNetworks are weight generator models, that use the learning signal from the target model to update the weight-generator~\citep{haHyperNetworks2017}. HyperNetworks have been successfully applied on multiple domains, model sizes, for regular training, architecture search, as well as meta-learning ~\citep{zhangGraphHyperNetworksNeural2019, knyazevParameterPredictionUnseen2021,deutschGeneratingNeuralNetworks2018,ratzlaffHyperGANGenerativeModel2019,zhmoginovHyperTransformerModelGeneration2022}.

Recent approaches have successfully leveraged self-supervised learning to build models that can be used for both these downstream tasks, without the need of explicitly labeled data. In particular, \citet{schurholtSelfSupervisedRepresentationLearning2021} use autoencoders (AEs) in weight space to learn neural network representations in a latent space --- which they name \textit{hyper-representations}. The encoder outputs latent representations of NN weights that can be used for discriminative downstream tasks, or fed to the decoder to reconstruct the model. Alternatively, the decoder can be fed synthetic latent representations to generate NN weights~\citep{schurholtHyperRepresentationsGenerativeModels2022}. 
While these \textit{hyper-representation} AEs show good predictive performance and low reconstruction mean-squared-error (MSE), they lack the fidelity to reconstruct or generate high performance models without further fine-tuning. This strongly limits the capability of such models for the generation of NN weights.

In this work, we investigate how adding a \textit{behavioral} component to the training loss of these autoencoders can help overcoming this limitation. Behavioral losses, where one uses \textit{queries} a model to analyze its outputs, have been successfully used in weight-space learning contexts, such as deep model alignment~\citep{navonEquivariantDeepWeight2024}, representation learning and model analysis~\cite{herrmannLearningUsefulRepresentations2024}, INR classification~\cite{deluigiDeepLearningImplicit2023}, etc. All of these works tend to indicate that behavior-based losses improve the performance of \textit{discriminative} downstream tasks; we show that they also lead to strongly improved performance with regard to \textit{reconstructive} and \textit{generative} downstream tasks. This finding will open up new opportunities in directly generating NN weights for a given task with a single forward pass.

\section{Structural Reconstruction Is Not Enough}
\label{sec:eigendecomposition}

\paragraph{Reconstruction in Weight Space Learning Prioritizes Coarse Features}

Previous work on representation learning of NN weights with AEs has largely focused on reconstructing weights in the structural sense, using the mean-squared-error (MSE)~\citep{schurholtSelfSupervisedRepresentationLearning2021}. These representations are effective for discriminative downstream tasks like model performance prediction. However, they struggle in reconstructing functional models without the need for further fine-tuning, even when minimizing the structural reconstruction error~\citep{schurholtHyperRepresentationsGenerativeModels2022, schurholtScalableVersatileWeight2024,soroDiffusionbasedNeuralNetwork2024a}. This difficulty may stem from at least two reasons. Firstly, there exist inductive biases inherent to undercomplete AEs, used in these weight representation learning approaches. Trained with the MSE loss, these AEs often exhibit a bias toward learning coarse, smoothed representations, as MSE minimizes reconstruction error by averaging over fine details~\citep{vincentStackedDenoisingAutoencoders2010}. Tasks that rely on high-resolution features, like texture recognition, further highlight the limitations of such biases \citep{zeilerVisualizingUnderstandingConvolutional2014}. This smoothing bias may explain the challenges in reconstructing functional models from weight representations. Secondly, measuring the distance between weights of NNs can be a poor metric of their (di)similarity due to the equivalence class of models as a result of permutations of the neurons or the weights \citep{entezari2021role}. In other words, this means that two models that are equivalent will be ``seen'' as different due to the sensitivity of the MSE loss to distance only. \looseness-1

\paragraph{Reconstruction Does Not Guarantee Generalization}
Recent work \cite{he2019asymmetric, dinh2017sharp} suggests that the solutions of NNs can be sensitive to local perturbations in the sense that there exist parameters close to the selected solution that have much worse performance than the selected solution. This implies that the reconstructed solution using the MSE loss can instantiate a worse performing model than the one it attempts to reconstruct: although the reconstructed solution is close to the original solution, it is in a region that has poor generalization performance. Furthermore, recent work by \citet{balestriero2024learning} shows that learning through reconstruction for image classification purposes is not enough to achieve good generalization performance due to the idiosyncratic aspects of the signal that the reconstruction task is unable to take into account. We therefore need further guidance to make the model reconstruct high-performing NNs. In the following, we incorporate a behavioral loss into our training objective, and assess how it impacts performance on a variety of downstream tasks. More particularly, we demonstrate that such a loss can strongly improve the performance of both reconstructed and generated model weights.\looseness-1

\section{Learning From Structure and Behavior}
\label{sec:methods}

Inspired by perceptual losses in computer vision \citep{esserTamingTransformersHighResolution2021b}, we propose a \emph{behavioral loss} that emphasizes the functional similarity between the original and reconstructed NNs. In the following, we first define the notation used throughout this paper, and then highlight the differences between the structural and behavioral losses.

\paragraph{Behavioral Loss $\mathcal{L}_B$}

Let:
$\mathbf{\theta}_j \in \Theta \subset \mathbb{R}^p$ be the parameters of the $j$-th NN from the model zoo, for $j = 1, \dots, k$;
$\hat{\theta}_j = g_w(\theta_j)$ be the reconstructed parameters via the AE with learnable parameters $w$;
$f_{\theta}: \mathcal{X} \rightarrow \mathcal{Y}$ denotes an NN that maps on $\mathcal{Y} \subset \mathbb{R}^{k}$ with weights $\theta$  ($f_{\theta_{j}}$ and $f_{\hat{\theta}_{j}}$ are the original and reconstructed NN with parameters $\theta_{j}$ and $\hat{\theta}_{j}$ respectively), and
$\{ x_i \}_{i=1}^n$ be a set of \textit{queries} of input samples from the sample space $\mathcal{X}$.
The empirical behavioral loss is defined as:
\begin{equation}
    \mathcal{L}_{B} = \frac{1}{2kn} \sum_{j=1}^k \sum_{i=1}^n \left\| f_{\hat{\theta}_j}(x_i) - f_{\theta_j}(x_i) \right\|^2
    \label{eq:behavioral_loss}
\end{equation}
This loss measures the discrepancy between the outputs of the original and reconstructed models over the queries, emphasizing functional equivalence. We illustrate the behavioral loss in Figure~\ref{fig:overview}.

\paragraph{Composite Representation Learning Loss}

\citet{schurholtSelfSupervisedRepresentationLearning2021} define a composite loss with two elements: $\mathcal{L}_C$ is the {contrastive loss}, promoting discriminative latent representations; $\mathcal{L}_S$ is the {structural loss}, measuring the parameter-wise difference between the original and reconstructed models. Their respective importance is modulated by a hyperparameter $\gamma$ as follows:

\begin{equation}
    \mathcal{L}_{C+S} = \gamma \mathcal{L}_C + (1 - \gamma) \mathcal{L}_S
    \label{eq:c_and_s_composite_loss}
\end{equation}

We integrate our behavioral loss directly into that composite loss. Both $\mathcal{L}_S$ and $\mathcal{L}_B$ compare the reconstructions to the original models, and as such, are \textit{reconstructive} losses. In our proposed composite loss, $\gamma$ modulates the weight of the \textit{contrastive} loss $\mathcal{L}_C$ with that of both \textit{reconstructive} losses. Within the \textit{reconstructive} losses, we define a new hyperparameter $\beta$ that modulates the weight of the \textit{structural} loss $\mathcal{L}_S$ with respect to the \textit{behavioral} loss $\mathcal{L}_B$.

\begin{equation}
    \mathcal{L} = \gamma \mathcal{L}_C + (1 - \gamma) \left( \beta \mathcal{L}_S + (1 - \beta) \mathcal{L}_B \right)
    \label{eq:composite_loss}
\end{equation}

\paragraph{Gradient Analysis} To understand how the behavioral loss influences the AE training differently from the structural loss, we analyze their gradients with respect to the AE parameters $w$.
The structural loss is realized as a mean squared error (MSE) over the parameters $\theta$:
\begin{equation}
    \mathcal{L}_{S} = \frac{1}{2k} \sum_{j=1}^k \left\| \hat{\theta}_j - \theta_j \right\|^2.
    \label{eq:structural_loss}
\end{equation}
Its gradient with respect to the AE's parameters $w$ is straightforward:
\begin{equation}
    \frac{\partial \mathcal{L}_{S}}{\partial w} = \frac{1}{k} \sum_{j=1}^k \Delta \theta_j ^\top \frac{\partial \hat{\theta}_j}{\partial w},
    \label{eq:grad_structural_loss}
\end{equation}
where $\Delta \theta_j = \hat{\theta}_j - \theta_j$. The gradient of the behavioral loss with respect to the AE parameters $w$ is:
\begin{equation}
    \frac{\partial \mathcal{L}_{B}}{\partial w} = \frac{1}{kn} \sum_{j=1}^k \sum_{i=1}^n \left( f_{\hat{\theta}_j}(x_i) - f_{\theta_j}(x_i) \right)^\top \frac{\partial f_{\hat{\theta}_j}(x_i)}{\partial \hat{\theta}_j} \frac{\partial \hat{\theta}_j}{\partial w},
    \label{eq:grad_behavioral_loss}
\end{equation}
where we abuse the partial derivative notation $\frac{\partial f_{\hat{\theta}_{j}}(x_{i})}{\partial \hat{\theta}_{j}}$ to mean taking the partial derivative with respect to $\theta$ evaluated at $\hat{\theta}_{j}$.

Assuming that the reconstructed parameters are close to the original ones (i.e., $\Delta \theta_j$ is small), which is a justified assumption when $||\hat{\theta}_{j} - {\theta}_{j}||_{2}^{2}$ is part of the loss function and the AE is sufficiently expressive, we can approximate $f_{\hat{\theta}_j}(x_i)$ using a first-order Taylor expansion around $\theta_j$:
\begin{equation}
    f_{\hat{\theta}_j}(x_i) \approx f_{\theta_j}(x_i) + J_{\theta_j}(x_i) \Delta \theta_j
    \label{eq:taylor_expansion}
\end{equation}
where $J_{\theta_j}(x_i) = \frac{\partial f_{\theta_j}(x_i)}{\partial \theta_j}$, with $J_{\theta_{j}}(x_i)^\top \in \mathbb{R}^{p\times k}$, is the Jacobian of the model output with respect to its parameters.
Substituting into the gradient:
\begin{align}
    \frac{\partial \mathcal{L}_{B}}{\partial w} &\approx \frac{1}{kn} \sum_{j=1}^k \sum_{i=1}^n \left( J_{\theta_j}(x_i) \Delta \theta_j \right)^\top J_{\hat{\theta}_j}(x_i) \frac{\partial \hat{\theta}_j}{\partial w} \nonumber \\
    &= \frac{1}{k} \sum_{j=1}^k \Delta \theta_j^\top \left( \frac{1}{n} \sum_{i=1}^n J_{\theta_j}(x_i)^\top J_{\hat{\theta}_j}(x_i) \right) \frac{\partial \hat{\theta}_j}{\partial w}.
    \label{eq:grad_behavioral_simplified}
\end{align}
This allows us to approximate the gradient of the behavioral loss as:
\begin{equation}
    \frac{\partial \mathcal{L}_{B}}{\partial w} \approx \frac{1}{k} \sum_{j=1}^k \Delta \theta_j^\top F_j \frac{\partial \hat{\theta}_j}{\partial w},
    \label{eq:grad_behavioral_final}
\end{equation}
where

\begin{equation}
    F_j = \frac{1}{n} \sum_{i=1}^n J_{\theta_j}(x_i)^\top J_{\hat{\theta}_j}(x_i).
\end{equation}

Comparing the behavioral loss gradient (Eq.~\ref{eq:grad_behavioral_final}) to the structural loss gradient (Eq.~\ref{eq:grad_structural_loss}) shows that both depend on $\Delta \theta_j$, the difference between the original and reconstructed parameters, and $\frac{\partial \hat{\theta}_{j}}{\partial w}$. However, while the structural loss gradient measures the linear alignment between $\Delta \theta_{j}$ and $\frac{\partial\hat{\theta}_{j}}{\partial w}$, the behavioral loss gradient measures the linear alignment between $\Delta \theta_{j}$ and a vector $F_{j}\frac{\partial \hat{\theta}_{j}}{\partial w}$. $F_{j}$ plays a key role in the gradients because it provides information on the average sensitivity of the original and reconstructed NNs to changes in the weights, as well as the average linear alignment between their corresponding gradients w.r.t. the weights.\looseness-1

\paragraph{The Choice of Queries Impacts the Learned Representations}
Two NNs trained on the same dataset with similar performance on the domain of the training data can have different behavior outside that domain. Similarly, since the behavioral loss depends on the queries used to compute it, the choice of those will influence which aspects of the behavior of the original models will be reconstructed. If the queries used for the behavioral loss come from a different domain than the one of the training data, the AE will attempt to match the behavior of the original and reconstructed model on parts of the domain where the NN from the zoo did not have training data, and where its behavior is therefore ill-defined. This implies that the reconstructed and original model will match performance on parts of the domain that is of no interest, while on the parts of interest the performance might be arbitrarily different. Hence, the choice of the queries plays an important role in the performance of the reconstructed models. In Appendix~\ref{ap:query_set} we explore the hypothesis that randomly generated queries lead to poor performing reconstructed models, while queries from the same or similar distribution as the training data lead to good performance of the reconstructed models.

\section{Experiments}
\label{sec:experiments}

In the following, we build on the conclusions of Section~\ref{sec:methods} and experimentally evaluate whether the inclusion of a behavioral element to the loss function leads to better performance on a selection of downstream tasks from the literature. We first describe our general experimental setup, and then compare different variants of our loss functions for each reconstructive and generative downstream tasks. Additionally, we show a small positive influence of the behavioral loss in Appendix~\ref{ap:discriminative}.

\subsection{Experimental Setup}
\label{sec:experimental_setup}

\paragraph{Model Zoos} In this work, we understand a model zoo as a structured population of models using the same architecture and trained on the same data. We use three different model zoos \cite{schurholt2022model} of convolutional neural networks (CNN), on the SVHN~\citep{SVHN}, CIFAR-10~\citep{CIFAR-10} and EuroSAT~\citep{helber2019eurosat} datasets. Every model zoo is built with the same grid of hyperparameters. In total, every zoo is composed of $1,200$ models, with $10,853$ parameters each, trained over $50$ epochs. We provide additional details about the model zoos generation, their architecture and their hyperparameters in Appendix~\ref{ap:model_zoos_generation}. The model zoos are randomly separated into disjoint train, validation and test splits with respective proportions $\{80\%, 5\%, 15\%\}$.

\paragraph{Hyper-Representation AEs} We use \texttt{SANE}, as the backbone to train hyper-representations as in  \citet{schurholtScalableVersatileWeight2024}, where the weights of the original model are tokenized, then fed to an encoder that generates one latent representation per token. A projection head is used for the contrastive loss $\mathcal{L}_C$, whereas all embeddings are fed to the decoder. With an original token length of $289$ and an embedded dimension of $64$, our compression ratio is $4.52$. While \texttt{SANE} allows the use of windows of tokens rather than entire models, we systematically feed and reconstruct an entire model at once; we discuss this limitation in more details in Section~\ref{sec:conclusion}. We train our hyper-representation models on the train split of the corresponding model zoo, using the checkpoints corresponding to training epochs $\{20, 30, 40, 50\}$. We provide additional details about hyper-representation AEs training in Appendix~\ref{ap:hr_training}. Additionally, we extend these results with some ablation experiments in Appendix~\ref{ap:ablations}.

\paragraph{Loss Functions} We use the composite loss function defined in Equation~\ref{eq:composite_loss}. It combines three elements: a contrastive loss in latent space $\mathcal{L}_C$, and two losses in reconstructed space, one structural $\mathcal{L}_S$, one behavioral $\mathcal{L}_B$. Their relative weights are controlled by two hyperparameters $\gamma, \beta \in [0, 1]$. When $\gamma = 0$ the contrastive loss is not used; when $\beta = 0$ the structural loss is not used; and when $\beta = 1$ the behavioral loss is not used. When using both the contrastive loss and a reconstructive loss, we set $\gamma = 0.05$, as in the original \texttt{SANE} implementation. When using both the structural and behavioral loss, we set $\beta = 0.1$. This value follows from an exploration of possible values on our validation set and has shown to perform best, as detailed in Appendix~\ref{ap:selecting_beta}. The case where $\gamma = 0.05$ and $\beta = 1$ corresponds to the existing \texttt{SANE} implementation and serves as our baseline. We implement the contrastive loss $\mathcal{L}_C$ the same way as in \texttt{SANE}, with NTXent~\citep{sohn2016ntxent} where the augmentations used are permutations of weights that do not alter the behavior of the NN. $\mathcal{L}_S$ and $\mathcal{L}_B$ are implemented using the mean-squared-error loss, which is equivalent to the $L^2$ loss up to a constant factor and a square root.
\paragraph{Queries for $\mathcal{L}_B$} Computing the behavioral loss $\mathcal{L}_B$ requires us to use some queries to feed to both the original and reconstructed models, so as to compare their outputs. For every batch, we sample $n_{queries} = 256$ images from the training set used to train the corresponding model zoo.

\subsection{Combining Structure and Behavior Increases Performance for Every Task}
\label{sec:exp_results}

\begin{figure*}[t!]
    \centering
    \includegraphics[width=0.9\textwidth]{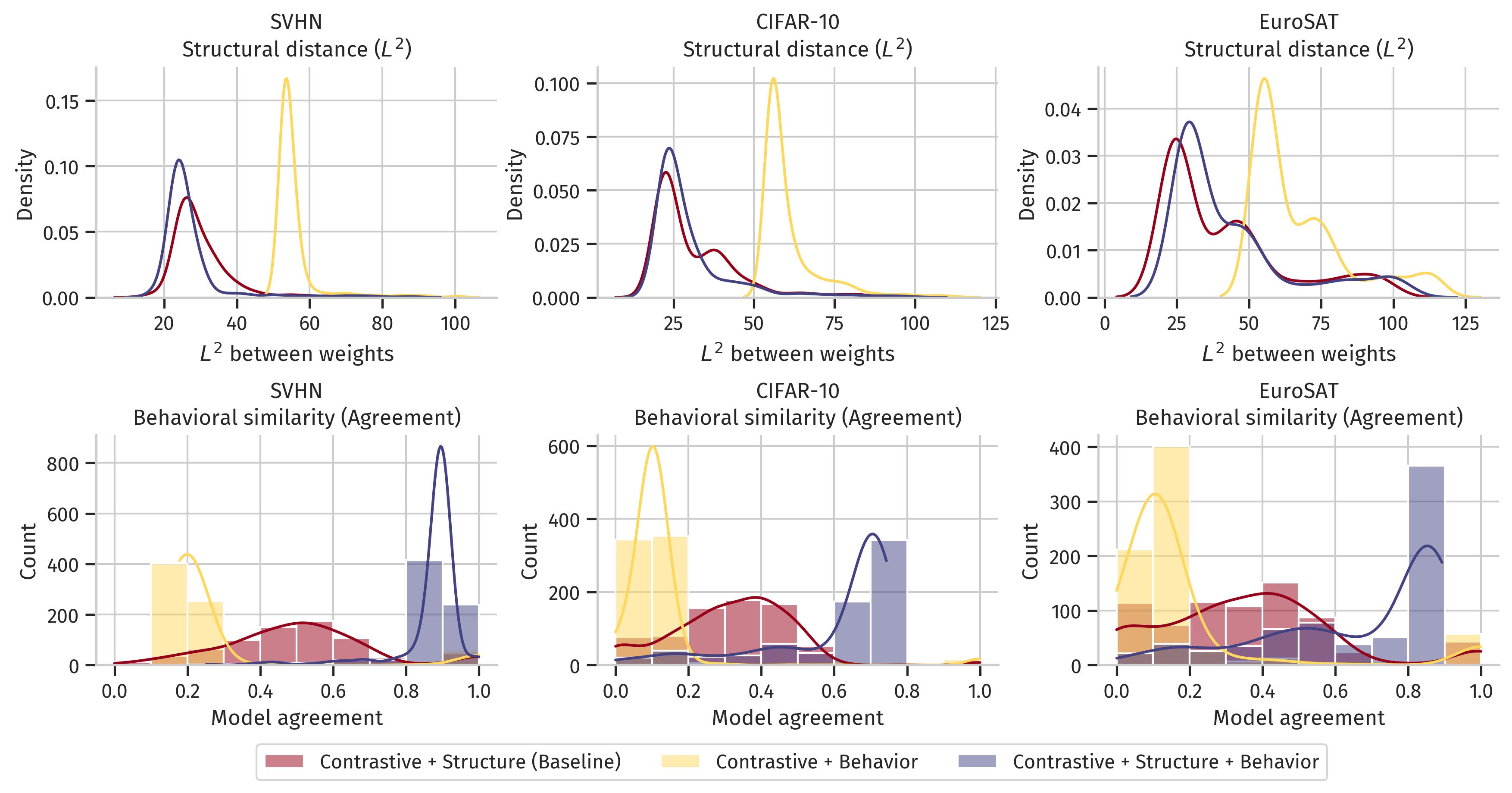}
    \caption{Evaluation of the reconstructive downstream tasks, pairwise between models from the model zoo and their reconstructions, depending on the losses used to train the hyper-representation AE. Each column represents one of our model zoos. \textbf{Top:} distribution of pairwise structural $L^2$ distances. \textbf{Bottom:} distribution of the pairwise behavioral similarities, measured with model agreement. On the structural side, we see that using the structural loss is sufficient to concentrate most pairwise distances around some low value. On the behavioral side, we see that using the behavioral loss only yields the worst performance, even compared to the structural loss. Using both the structural and behavioral losses is necessary to achieve high levels of agreement. Most models show high levels of agreements, but since a few show low levels of agreement the standard deviation shown in Table~\ref{tab:reconstructive_dstk} can be high.}
    \label{fig:reconstructive_dstk_pairwise}
\end{figure*}

\subsubsection{Reconstructive Downstream Tasks} 

We aim for the reconstructed models to be similar to the originals, both in structure and behavior. Our model zoos are diverse, and contain both high- and low-performing models: we expect a poorly performing model to be reconstructed as a poorly performing model, and conversely for well-performing models. We evaluate the fidelity of reconstructed models first pairwise, then in distribution over the whole test split. All metrics are computed over the test split of the model zoos, which only includes models not seen by the hyper-representation AE at training time. We show results for pairwise evaluation in Figure~\ref{fig:reconstructive_dstk_pairwise}, and report more detailed results in Appendix Table~\ref{tab:reconstructive_dstk}.\looseness-1

We first assess the structural reconstruction fidelity by comparing the pairwise $L^2$ distances between a model weights and its reconstruction. We note that using $\mathcal{L}_S$ is sufficient and necessary to get low structural reconstruction errors, with large differences between models that use it and those that do not. We then measure the behavioral distance between models and their reconstruction using model agreement. Conversely to the structural distance, we note that using the behavioral loss does not guarantee high model agreement: models trained with $\mathcal{L}_B$ but without $\mathcal{L}_S$ show both the highest structural error and lowest model agreement. At the same time, models trained with both $\mathcal{L}_S$ and $\mathcal{L}_B$ show much higher agreement than all others. In particular, when comparing those models to our baseline that uses $\mathcal{L}_C$ and $\mathcal{L}_S$, we see that including $\mathcal{L}_B$ is essential to achieve a reconstruction that is behaviorally similar to the original model.\looseness-1

\begin{figure*}[t]
    \centering
    \includegraphics[width=0.9\textwidth]{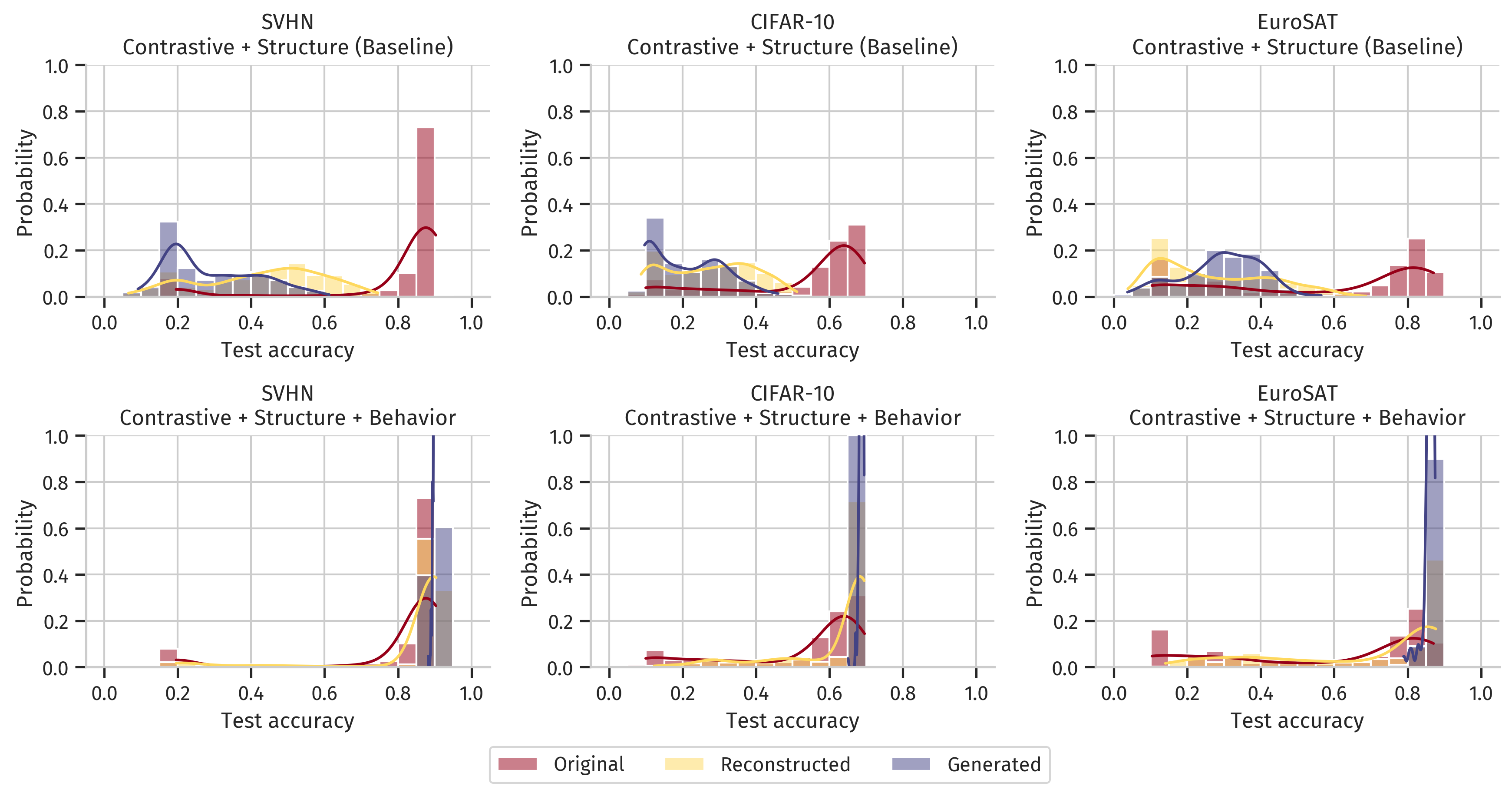}
    \caption{Evaluation of the reconstructive and generative downstream tasks, shown as distributions of the test accuracy of different models, depending on whether they are part of the original model zoo, reconstructions of models from that model zoo, or generated models. Each column represents one of our model zoos, while the row show what loss has been used to train the specific hyper-representation model. The top row shows results for our baseline, that uses the contrastive $\mathcal{L}_C$ and structural $\mathcal{L}_S$ losses. The bottom row represents our hyper-representation AEs, which in addition are also trained with a behavioral loss $\mathcal{L}_S$. We note that for the baseline, neither the reconstructed nor the generated models can match the performance of the original models. On the other hand, when adding a behavioral element to the loss, they match the performance of the most accurate models from the original zoo.}
    \label{fig:generative_dstk}
\end{figure*}

We compare the distributions of model accuracies for models from the zoos and their reconstructions in Figure~\ref{fig:generative_dstk}. We compare the baseline AE that uses $\mathcal{L}_C$ and $\mathcal{L}_S$ only to our model that is also trained with $\mathcal{L}_B$. In all cases, the baseline fails to reconstruct models of performance comparable to the ones in the original zoo. On the other hand, when including $\mathcal{L}_B$, reconstructed models are very performant --- at the same time, there seems to be a bias towards reconstructing models with higher performance than the original. This could be caused by an over-representation of good performing models in our model zoos, compared to poorly performing ones.

\subsubsection{Generative Downstream Tasks}

In addition to reconstructing exisiting models, fixing the hyper-representation AE's ability to output performant model weights opens up the way for generating new, unseen model weights. In their existing work, \citet{schurholtHyperRepresentationsGenerativeModels2022} generate models using their hyper-representations, but their performance is limited and they have to be re-trained for a few epochs to achieve performance comparable to models in the original zoo. In the following, we evaluate whether hyper-representation AEs trained with a behavioral loss can perform better at model weight generation.

To generate model weights, we first need to generate corresponding synthetic latent representations. To do so, we select anchor models from the model zoo that are themselves well-performing (more details in Appendix~\ref{ap:hr_training}) and compute their latent representations. With the goal of mitigating the curse of dimensionality, we then use PCA to project those in a lower dimensional space of size $32$, and use a kernel density estimate (KDE) to model the distribution of the projected embeddings. For the KDE, we further assume that the coordinates are orthogonal to each other. We generate new data points by sampling from that KDE, and inverse project back into the latent representation space. Finally, we feed these synthetic hyper-representations to the decoder and evaluate the resulting models.
We show results in Figure~\ref{fig:generative_dstk}.

We first note that when using the baseline hyper-representation AE, the generated models tend to be as good as or slightly worse than the reconstructed models. This level of performance is low compared to the anchors we use for generation, which for all three datasets have more than $60\%$ test accuracy. On the other hand, when adding our behavioral element to the loss function, we see that the generated models have a very high test accuracy, very close to the best models in the model zoo, as reported in Table~\ref{tab:gen_models_max_perf}. Their performance seem to match that of the best reconstructed models. Finally, we show in Appendix Table~\ref{tab:gen_models_diversity} that generated models are somewhat diverse, although less than the original anchors. This confirms that we do not generate identical models. 


\section{Related Work}

\paragraph{Weight Space Representation Learning}
Recent work in representation learning on NN weights has led to various approaches for analyzing and generating models weights. Hyper-Representations \citep{schurholtSelfSupervisedRepresentationLearning2021,schurholtHyperRepresentationsPreTrainingTransfer2022,schurholtHyperRepresentationsGenerativeModels2022,soroDiffusionbasedNeuralNetwork2024a} use an encoder-decoder architecture with contrastive guidance to learn weight representations for property prediction and model generation. 
Using a different learning task for weight generation, other methods employ diffusion on weights~\citep{peeblesLearningLearnGenerative2022, wangNeuralNetworkParameter2024,jinConditionalLoRAParameter2024a}. 
Our work differs by focusing on understanding and mitigating the inductive biases in weight space learning, particularly for AE-based approaches. 
Similar to our behavioral loss, HyperNetworks use the learning feedback from the target models to generate their weights \citep{haHyperNetworks2017,knyazevParameterPredictionUnseen2021,knyazevCanWeScale2023}. 
Graph representation methods \citep{zhangGraphHyperNetworksNeural2019,kofinasGraphNeuralNetworks2023,limGraphMetanetworksProcessing2023}, Neural Functionals \citep{zhouPermutationEquivariantNeural2023,zhouNeuralFunctionalTransformers2023,zhouUniversalNeuralFunctionals2024} and related approaches like Deep Weight Space (DWS)\citep{navonEquivariantArchitecturesLearning2023, zhangNeuralNetworksAre2023} learn equivariant or invariant representations of weights.
While these methods incorporate geometric priors of the weight space in encoder or decoder models, we focus in this work on AE-based approaches, as they cover the largest breadth of downstream tasks. Augmentations specific to weight space learning have also been developed~\citep{shamsianImprovedGeneralizationWeight2024}. As mentioned in Section~\ref{sec:experimental_setup}, our experiments focus on augmentations that do not alter the behavior of the NN, and therefore do not include such augmentations.

\begin{table*}[t]
    \caption{Maximum performance of selected models. `\texttt{Zoo}' describes models from the original model zoo; `\texttt{Recon.}' includes the reconstructions of the models from the test split of the model zoo; `\texttt{Gener.}' represents the models synthetically generated as described in Section~\ref{sec:exp_results}. `\texttt{$\Delta_{Acc}$ Rec.}' and `\texttt{$\Delta_{Acc}$ Gen.}' refer to the difference in performance between the best models in respectively the reconstructed and generated models on one side, and the original models on the other side. For the baseline that only uses the contrastive $\mathcal{L}_C$ and structural $\mathcal{L}_S$ losses, reconstruction and generation do not manage to build models of comparable performance with the original zoo. Conversely, when adding the behavioral loss $\mathcal{L}_B$, we achieve a maximum performance very close to that of the original zoo.}
    \label{tab:gen_models_max_perf}
    \begin{center}
    \tabcolsep=0.10cm
    \begin{tabular}{llccccc}
    & \textsc{\textbf{Losses}} & \textsc{\textbf{Zoo}} & \textsc{\textbf{Recon.}} & \textsc{\textbf{$\Delta_{Acc}$ Rec.}} & \textsc{\textbf{Gener.}} & \textsc{\textbf{$\Delta_{Acc}$ Gen.}} \\
     \midrule
    \textsc{SVHN} & $\mathcal{L}_C \oplus \mathcal{L}_S$ \textsc{\small{(Baseline)}} & 91.0\% & 74.5\% & -16.5\% & 61.3\% & -29.7\% \\
     & $\mathcal{L}_C \oplus \mathcal{L}_S \oplus \mathcal{L}_B$ & 91.0\% & 90.4\% & \textbf{-0.6\%} & 90.4\% & \textbf{-0.6\%} \\
     \midrule
    \textsc{CIFAR-10} & $\mathcal{L}_C \oplus \mathcal{L}_S$ \textsc{\small{(Baseline)}} & 70.1\% & 51.2\% & -18.9\% & 46.0\% & -24.1\% \\
     & $\mathcal{L}_C \oplus \mathcal{L}_S \oplus \mathcal{L}_B$ & 70.1\% & 69.5\% & \textbf{-0.6\%} & 69.5\% & \textbf{-0.6\%} \\
     \midrule
    \textsc{EuroSAT} & $\mathcal{L}_C \oplus \mathcal{L}_S$ \textsc{\small{(Baseline)}} & 88.5\% & 68.6\% & -19.9\% & 56.5\% & -32.0\% \\
     & $\mathcal{L}_C \oplus \mathcal{L}_S \oplus \mathcal{L}_B$ & 88.5\% & 87.7\% & \textbf{-0.8\%} & 87.5\% & \textbf{-1.0\%} 
    \end{tabular}
    \end{center}
\end{table*}

\paragraph{Probing-Based Losses in Weight Space} Other works have leveraged losses that are based on the response of models given a set of queries, first of which are HyperNetworks~\citep{haHyperNetworks2017}. \citet{deluigiDeepLearningImplicit2023} use weight-space autoencoders, but they feed the queries directly to the decoder which is then trained on the true labels. \citet{navonEquivariantArchitecturesLearning2023} propose an architecture that can be used in many different setups, one of them being domain adaptation, where they use the loss on the new domain. More closely related to our work, \citet{herrmannLearningUsefulRepresentations2024} leverage behavioral losses (which they name functionalist) to analyze recurrent neural networks (RNNs) with high performance. Their work differs from ours as they use the decoder as an emulator of the function represented by the original model, whereas we directly reconstruct the original model. \citet{navonEquivariantDeepWeight2024} use a composite loss in their deep weight space alignment experiment setup, with elements closely related to structure and behavior but tailored to their particular use-case.

\section{Discussion}
\label{sec:conclusion}

\paragraph{Limitations} Scaling tranformer-based architectures to large sequences is challenging, the present exploration is therefore limited to smaller models. We defer the implementation of a behavioral loss on larger NNs to future work. Another limitation is the computational overhead, as we need to reconstruct and test generated models at each training step. In Appendix~\ref{ap:ablation_compute} however, we show that for comparable computing time, using structure and behavior still outperforms fully-structural approaches.\looseness-1

\paragraph{Conclusion} Our work shows a strong synergy between structural and behavioral signals when training autoencoders on populations of trained NNs. We build a behavioral loss function to guide the learning of weight-space AEs towards the other features that are essential to reconstruct high-performing models, and theoretically explore how it differs from a purely structural loss. We demonstrate experimentally that adding a behavioral loss outperforms the purely structural baseline for discriminative, reconstructive as well as generative downstream tasks. More generally, our work shows that training self-supervised models in weight space requires a balance between structural and behavioral features to perform well. Our analysis uncovers hitherto unknown insights and opens up exciting research opportunities in the domain of weight-space representation learning.

\newpage

\subsubsection*{Code availability}

We make our code available at \href{https://doi.org/10.5281/zenodo.15051578}{doi.org/10.5281/zenodo.15051578}

\subsubsection*{Acknowledgments}

This work was funded by Swiss National Science Foundation research project grant 10001118. The LMU authors acknowledge support from the Elite Network Bavaria as well as the Munich Center for Machine Learning (MCML).

\bibliography{./iclr2025_conference}
\bibliographystyle{iclr2025_conference}

\newpage

\appendix

\section{Model zoos generation}
\label{ap:model_zoos_generation}

All the models in our zoos have the same architecture, described in Table~\ref{tab:zoo_arch}. For the EuroSAT dataset, we first resize the images to $32 \times 32$ so that for all three datasets, inputs are of dimension $32 \times 32 \times 3$ and outputs of dimension $10$. For all datasets, we use standardization with the mean and standard deviations for ImageNet. We do not use augmentations.

\begin{table}[h!]
    \caption{CNN architecture used for our experiments.}
    \label{tab:zoo_arch}
    \begin{center}
    \begin{tabular}{lcc}
    \multicolumn{1}{c}{\textsc{\textbf{Layer}}}  & \textsc{\textbf{Hyperparameter}} &  \textsc{\textbf{Value}} \\
    \midrule 
    Convolutional 1 & Channels in & 3 \\
    & Channels out & 16 \\
    & Kernel size & 3 \\
    \midrule
    MaxPool 1 & Kernel size & 2\\
    & Stride & 2\\
    \midrule
    ReLU 1 &  & \\
    \midrule
    
    Convolutional 2 & Channels in & 16 \\
    & Channels out & 32 \\
    & Kernel size & 3 \\
    \midrule
    MaxPool 2 & Kernel size & 2\\
    & Stride & 2\\
    \midrule
    ReLU 2 &  & \\
    \midrule
    
    Convolutional 3 & Channels in & 32 \\
    & Channels out & 15 \\
    & Kernel size & 3 \\
    \midrule
    MaxPool 3 & Kernel size & 2\\
    & Stride & 2\\
    \midrule
    ReLU 3 &  & \\
    \midrule
    
    Flatten & &\\
    \midrule
    
    Linear 1 & Dimension in & 60 \\
    & Dimension out & 20 \\
    \midrule
    ReLU 4 & &\\
    \midrule
    Linear 2 & Dimension in & 20 \\
    & Dimension out & 10 \\
    
    \end{tabular}
    \end{center}
\end{table}

\begin{figure}[ht!]
    \centering
    \includegraphics[width=\textwidth]{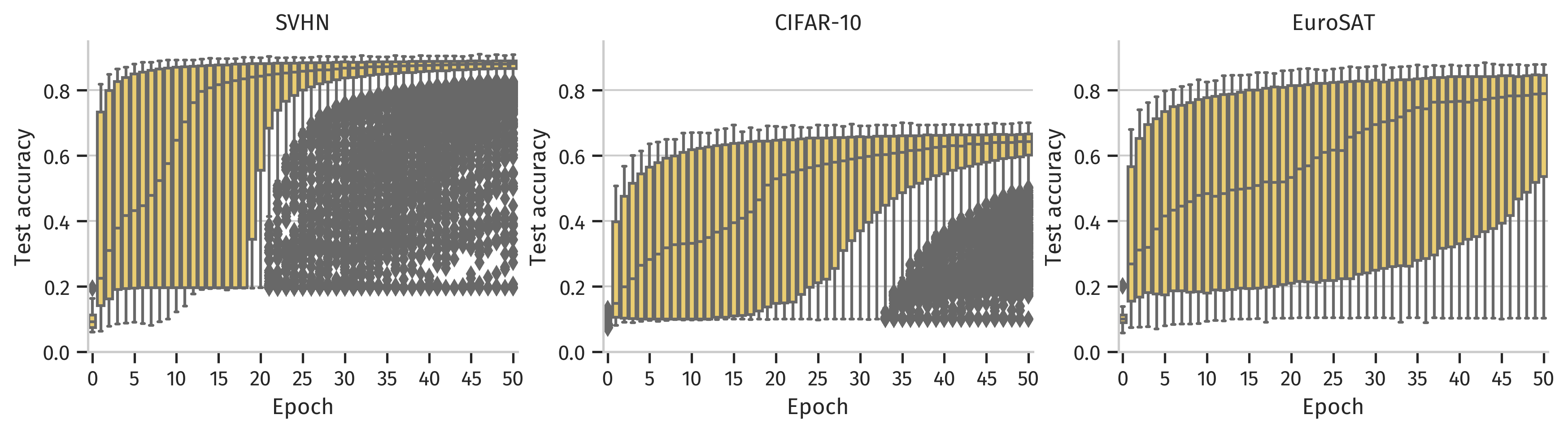}
    \caption{Distribution of the model test accuracies in our three model zoos, by training epoch.}
    \label{fig:model_zoos_acc}
\end{figure}

Models are trained for 50 epochs, with a batch size of 32. We use the Adam optimizer. To generate diverse models within our zoo, we vary other hyperparameters, as described in Table~\ref{tab:zoo_hyperparams}. This results in a total of $1,200$ models per zoo, with one checkpoint for each training epoch. In Figure~\ref{fig:model_zoos_acc}, we show the distribution of model test accuracies in all three zoos, throughout the training process.

\begin{table}[h!]
    \caption{Training hyperparameters for our model zoos. Kaiming initializations refer to work by \citet{heDelvingDeepRectifiers2015}.}
    \label{tab:zoo_hyperparams}
    \begin{center}
    \begin{tabular}{ll}
    \multicolumn{1}{c}{\textsc{\textbf{Hyperparameter}}}  &\multicolumn{1}{c}{\textsc{\textbf{Values}}} \\
    \midrule \\
    Initialization &Uniform, Normal, Kaiming Uniform, Kaiming Normal \\
    Learning rate & $1e-4$, $1e-4$, $2.5e-4$, $5e-4$, $7.5e-4$, $1e-3$ \\
    Weight decay &$1e-4$, $5e-4$, $1e-3$\\
    Seed &$0, 1, ..., 19$\\
    \end{tabular}
    \end{center}
\end{table}

\newpage
\section{Hyper-representation model training}
\label{ap:hr_training}

\paragraph{Pre-processing} While the original \texttt{SANE} implementation uses several pre-processing steps such as weight alignment with Git Re-Basin~\citep{ainsworth2022git} or standardization, we find empirically that such steps are not necessary anymore when using $\mathcal{L}_B$. We therefore only use a single pre-processing step, which is the pre-computation of behaviorally equivalent weight permutations of models in our zoo, which are used as augmentations and in the context of the contrastive loss $\mathcal{L}_C$.  

\paragraph{Hyperparameters} We train our hyper-representation transformer encoder and decoder with the following hyper-parameters: an input dimension of $289$, an output dimension of $64$, and a model dimension (\texttt{d\_model}) of $256$. The encoder was configured with $8$ attention heads (\texttt{num\_heads}) and comprised $8$ layers (\texttt{num\_encoder\_layers}). We use a batch size of $64$, and a weight decay of $3e-9$. After grid optimization, we select a learning rate of $1e-4$ for our baseline ($\mathcal{L}_C + \mathcal{L}_S$), and $1e-5$ for all others (i.e., those that use $\mathcal{L}_B$). Training is done for $100$ epochs.

\paragraph{Anchors for the generative downstream tasks} Regarding the selection of anchor models for the generative downstream task, we select only models with good performance compared to the rest of the zoo. As shown in Figure~\ref{fig:model_zoos_acc}, the distribution of accuracies varies a lot depending on the zoo we take into consideration. For this reason, we choose different test accuracy thresholds for each zoo: $80\%$ for SVHN, $60\%$ for CIFAR-10 and $70\%$ for EuroSAT.

\paragraph{Computational load} We ran all experiments on NVIDIA Tesla V100 GPU. We train for a total of $100$ epochs, and on a training set composed of $4,080$ checkpoints ($1,020$ different models, training epochs $\{20, 30, 40, 50\}$). When the behavioral loss is not used, training takes around $4,400$ seconds or a little less than 1 hour and 15 minutes, while when using the behavioral loss, training takes around $8,300$ seconds or a little more than 2 hours and 15 minutes. There is a notable difference because instead of only comparing tokens to tokens, the reconstructed tokens have to be converted back into a usable model, and then a forward pass must be done over the reconstructed model. Fortunately, computing this for each model in a batch is parallelized, hence why the difference in computation time is still under a factor of $2$. While indeed more expensive than the fully-structural approach, we deem the trade-off for increased performance acceptable.

\newpage
\section{Additional results}

In this Section, we show additional results in Table format that complement those shown in Section~\ref{sec:exp_results}. In Table~\ref{tab:reconstructive_dstk}, we show structural distance and behavioral similarity between models and their reconstructions, pairwise. In Table~\ref{tab:gen_models_diversity}, we study the diversity of the generated models, both in terms of structure and behavior. Finally, in Section~\ref{ap:discriminative}, we show that adding a behavioral element to our loss also has a small positive influence on discriminative downstream tasks.

\begin{table}[ht]
    \caption{Reconstructive downstream tasks performance. We evaluate structural reconstruction with the average $L^2$ distances between the weights of test split models and their reconstructions. We evaluate behavioral reconstruction with the average classification agreement between test split models and their reconstructions. Standard deviation is indicated between parentheses. \textsc{Bl} indicates the baseline. While using the structural loss $\mathcal{L}_S$ is sufficient and necessary to get low strucural reconstruction distance, the behavioral loss $\mathcal{L}_B$ alone does not suffice to reconstruct behaviorally close models. Only the combination of $\mathcal{L}_S$ and $\mathcal{L}_B$ allows the reconstruction of models that are close both in terms of structure and behavior.}
    \label{tab:reconstructive_dstk}
    \begin{center}
    \tabcolsep=0.05cm
    \begin{tabular}{lcccccc}
    \multicolumn{1}{c}{\textsc{\textbf{Losses}}} & \multicolumn{3}{c}{\textsc{\textbf{Structure ($L^2$ distance)}}} & \multicolumn{3}{c}{\textbf{\textsc{Behavior (Agreement)}}}\\
    & \textsc{SVHN} & \textsc{CIFAR-10} & \textsc{EuroSAT} & \textsc{SVHN} & \textsc{CIFAR-10} & \textsc{EuroSAT}\\
    \midrule 
    $\mathcal{L}_C \oplus \mathcal{L}_S$ \textsc{\small{(Bl)}} & 30.7 (±10) &  31.3 (±13) &  \textbf{41.4 (±22)} & 50.0\% (±20\%) &  31.9\% (±16\%) &  35.9\% (±24\%) \\
    \midrule
    $\mathcal{L}_B$ & 57.8 (±6) &   59.5 (±9) &  67.5 (±17) & 26.6\% (±22\%) &   10.3\% (±8\%) &  19.2\% (±25\%) \\
    $\mathcal{L}_C \oplus \mathcal{L}_B$ & 55.7 (±7) &   60.7 (±9) &  67.4 (±17) & 26.6\% (±22\%) &  11.8\% (±14\%) &  18.5\% (±25\%) \\
    $\mathcal{L}_S \oplus \mathcal{L}_B$ & \textbf{26.9 (±10)} &  \textbf{29.6 (±13)} &  45.5 (±21) & 86.1\% (±11\%) &  59.4\% (±20\%) &  \textbf{66.5\% (±25\%)} \\
    $\mathcal{L}_C \oplus \mathcal{L}_S \oplus \mathcal{L}_B$ & 27.1 (±10) &  30.0 (±13) &  43.6 (±21) & \textbf{87.0\% (±9\%)} &  \textbf{59.6\% (±19\%)} &  65.8\% (±25\%)
    \end{tabular}
    \end{center}
\end{table}

\begin{table}[ht]
    \caption{Diversity of generated models, expressed as the mean pairwise $L^2$ distance between all models in the selected set, computed either on weights (structure) or predictions (behavior). Anchors are the models from the model zoo used as a basis to generate synthetic latent representations. In all cases, there is some diversity in then generated models, showing that we do not generate identical models. However, when using the behavioral loss $\mathcal{L}_B$, we see that both the structural and behavioral diversity of generated models is lower.}
    \label{tab:gen_models_diversity}
    \begin{center}
    \tabcolsep=0.05cm
    \begin{tabular}{lcccccc}
    \multicolumn{1}{c}{\textsc{\textbf{Losses}}} & \multicolumn{3}{c}{\textsc{\textbf{Structure ($L^2$ distance)}}} & \multicolumn{3}{c}{\textbf{\textsc{Behavior ($L^2$ distance)}}}\\
    & \textsc{SVHN} & \textsc{CIFAR-10} & \textsc{EuroSAT} & \textsc{SVHN} & \textsc{CIFAR-10} & \textsc{EuroSAT}\\
    \midrule 
    \textsc{Anchors} & 23.3 (±4) & 22.9 (±4) & 23.0 (±5) & 3.6 (±0) & 4.7 (±0) & 4.2 (±1) \\
    \midrule
    $\mathcal{L}_C \oplus \mathcal{L}_S$ \textsc{\small{(Baseline)}} & 13.4 (±3) & 14.9 (±4) & 16.8 (±5) & 4.1 (±2) & 4.9 (±2) & 8.7 (±2) \\
    \midrule
    $\mathcal{L}_B$ & 1.6 (±0) & 3.5 (±1) & 6.3 (±1) & 0.0 (±0) & 0.2 (±0) & 0.1 (±0) \\
    $\mathcal{L}_C \oplus \mathcal{L}_B$ & 1.7 (±0) & 1.5 (±0) & 3.0 (±1) & 0.0 (±0) & 0.1 (±0) & 0.1 (±0) \\
    $\mathcal{L}_S \oplus \mathcal{L}_B$ & 1.6 (±0) & 1.5 (±0) & 2.8 (±1) & 0.6 (±0) & 0.8 (±0) & 1.2 (±1) \\
    $\mathcal{L}_C \oplus \mathcal{L}_S \oplus \mathcal{L}_B$ & 1.5 (±0) & 1.3 (±0) & 2.1 (±1) & 0.7 (±0) & 0.7 (±0) & 1.1 (±1)
    \end{tabular}
    \end{center}
\end{table}

\subsection{Discriminative Downstream Tasks} 
\label{ap:discriminative}

We consider two different discriminative downstream tasks, where we try to predict either a model's test accuracy or its generalization gap (defined as the difference between the train and test accuracies). To do so, we compute its hyper-representation using AEs trained with different elements of the composite loss. Following existing evaluation setups from the literature, we average all embedded tokens together into a $64$ dimensional ``center of gravity'' of the embeddings, and use a linear probe on top of it. The probe is trained on the models in the train split, and evaluated on the held-out test split. We show results in Table~\ref{tab:discriminative_dstk}. \newpage

\begin{table*}[ht]
    \caption{Discriminative downstream tasks performance. We predict the test accuracy and generalization gap of our models based on their latent representation, using a linear probe. We give the $R^2$ score for predictions on the held-out test split. We express the `Improvement' as the $R^2$ score of $\mathcal{L}_C \oplus \mathcal{L}_S \oplus \mathcal{L}_B$ minus the score of the baseline $\mathcal{L}_C \oplus \mathcal{L}_S$. In all cases, the most performant loss combination includes both the structural loss $\mathcal{L}_S$ and the behavioral loss $\mathcal{L}_B$.}
    \label{tab:discriminative_dstk}
    \begin{center}
    \tabcolsep=0.10cm
    \begin{tabular}{lcccccc}
    \multicolumn{1}{c}{\textsc{\textbf{Losses}}} & \multicolumn{3}{c}{\textsc{\textbf{Test accuracy}}} & \multicolumn{3}{c}{\textbf{\textsc{Generalization gap}}}\\
    & \textsc{SVHN} & \textsc{CIFAR-10} & \textsc{EuroSAT~~~~~~~~} & \textsc{SVHN} & \textsc{CIFAR-10} & \textsc{EuroSAT}\\
    \midrule
    $\mathcal{L}_C \oplus \mathcal{L}_S$ \textsc{\small{(Baseline)~~~~~~}} & 0.742 & 0.890 & 0.957 & 0.347 & 0.700 & 0.465 \\
    \midrule
    $\mathcal{L}_B$ & 0.538 & 0.771 & 0.901 & 0.286 & 0.576 & 0.296 \\
    $\mathcal{L}_C \oplus \mathcal{L}_B$ & 0.752 & 0.893 & 0.950 & 0.337 & 0.710 & 0.482 \\
    $\mathcal{L}_S \oplus \mathcal{L}_B$ & \textbf{0.887} & 0.939 & 0.966 & \textbf{0.378} & \textbf{0.785} & 0.484 \\
    $\mathcal{L}_C \oplus \mathcal{L}_S \oplus \mathcal{L}_B$ & 0.886 & \textbf{0.947} & \textbf{0.969} & 0.368 & \textbf{0.785} & \textbf{0.529} \\
    \midrule
    \textsc{Improvement} & 0.144 & 0.056 & 0.012 & 0.021 & 0.085 & 0.063 \\
    \end{tabular}
    \end{center}
\end{table*}

Although we built mostly to improve reconstructive accuracy, our behavioral loss also improves the performance of our discriminative downstream tasks: in all cases, the most performant loss combination includes both $\mathcal{L}_S$ and $\mathcal{L}_B$. Since we use linear probes, that means that the behavioral loss plays a part in better structuring the latent space, and that combining it with the structural loss synergizes well. We interpret that the structural loss serves as a regularization of the behavioral loss, meaning that the behaviorally optimal weights are not far from the original weights in an $L^2$ sense.\looseness-1

\newpage
\section{Ablation experiments}
\label{ap:ablations}

\subsection{Contrastive loss is still useful}

\begin{figure}[ht]
    \centering
    \includegraphics[width=\textwidth]{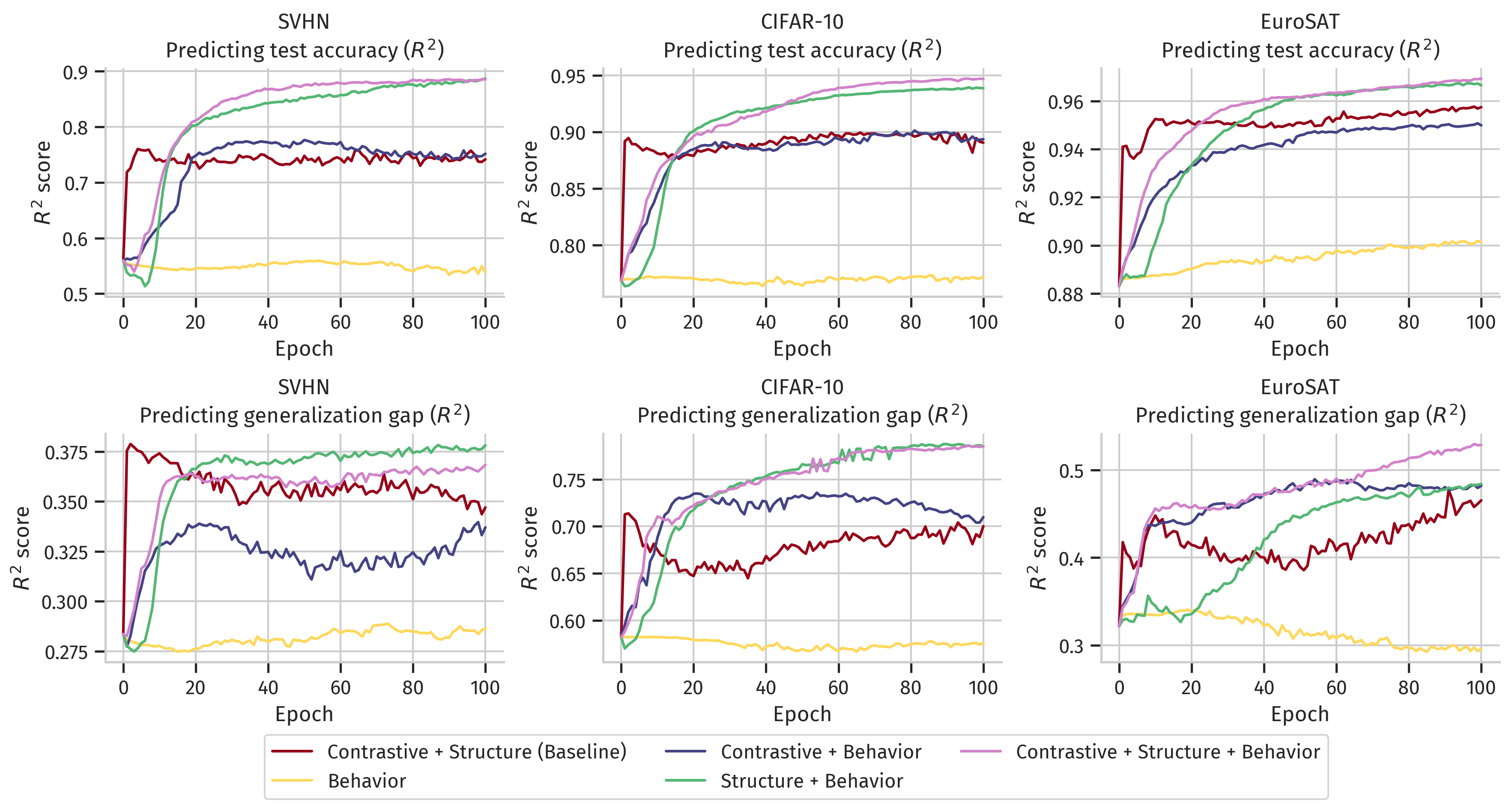}
    \caption{Comparison of the performance of the discriminative downstream tasks by training epoch of \texttt{SANE}. We first note that both models that use $\mathcal{L}_S$ and $\mathcal{L}_B$, show a more stable increase in performance with each epoch, and mostly outperform other models. Then, comparing the model that uses $\mathcal{L}_S \oplus \mathcal{L}_B$ with the one that uses $\mathcal{L}_C \oplus \mathcal{L}_S \oplus \mathcal{L}_B$, we qualitatively note that in most cases, performance grows faster with the number of epochs.}
    \label{fig:discriminative_dstk}
\end{figure}

In Section~\ref{sec:exp_results}, we show that when combining the structural $\mathcal{L}_S$ and behavioral $\mathcal{L}_B$ losses, the performance of all downstream tasks is improved. This raises an additional question: is using the contrastive loss $\mathcal{L}_C$ still relevant? The results shown in Tables~\ref{tab:discriminative_dstk} and \ref{tab:reconstructive_dstk} look inconclusive in that regard. As $\mathcal{L}_C$ focuses in structuring the latent space and has little influence over the decoder, we mostly evaluate it with regard to the discriminative downstream tasks that take place in the latent space, in Figure~\ref{fig:discriminative_dstk}.

We  compare both models that use $\mathcal{L}_S$ and $\mathcal{L}_B$: the one that uses $\mathcal{L}_C$ and the one that does not. Qualitatively, the performance of the model that uses $\mathcal{L}_C$ grows faster with the number of training epochs compared to that of the model that does not use $\mathcal{L}_C$. Additionally, we note from Table~\ref{tab:discriminative_dstk} that when predicting the generalization gap for the EuroSAT zoo, the model that uses $\mathcal{L}_C$ outperforms the one that does not by $0.045$, whereas the largest difference in performance in the other direction happens for predicting the generalization gap for the SVHN zoo, and is only of $0.01$, or $4.5$ times inferior. Although relatively weak, empirical evidence seems to indicate that using the contrastive loss $\mathcal{L}_C$ together with both structure $\mathcal{L}_S$ and behavior $\mathcal{L}_B$ remains relevant.

\subsection{MSE Is the Most Stable Behavioral Loss}

\begin{figure}[ht]
    \centering
    \includegraphics[width=\textwidth]{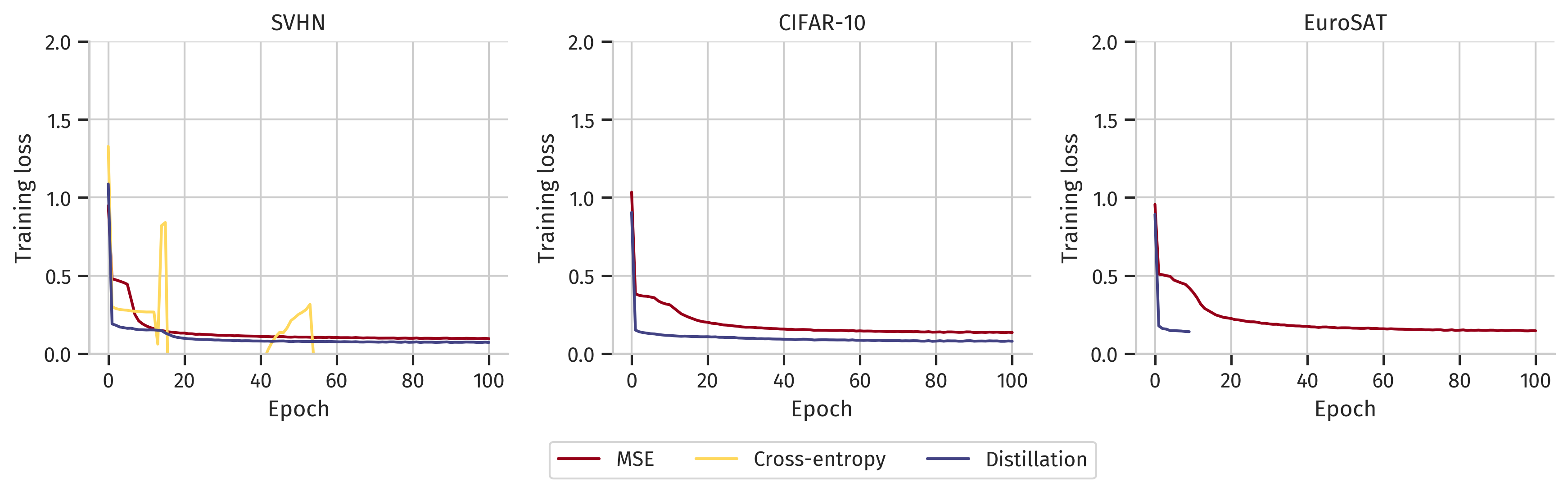}
    \caption{Training loss of hyper-representation models that are trained with all three of contrastive $\mathcal{L}_C$, structural $\mathcal{L}_S$ and behavioral $\mathcal{L}_B$ elements, but where we vary how the behavioral loss is computed. We note that only the MSE loss yields stable results across all model zoos.}
    \label{fig:train_loss_different_functions}
\end{figure}

In Section~\ref{sec:experimental_setup}, we use a MSE loss (which is equivalent to the $L^2$ distance up to a constant factor and a square root) over the predictions of a model and that of its reconstruction. There are, however, other losses that could be used in that context. We explore those empirically in this Section. In particular, we tested using either a cross-entropy loss, or a distillation loss~\citep{hinton2015distilling} with a temperature of $2$. We explore these on hyper-representations trained with all three elements of the composite loss: contrastive $\mathcal{L}_C$, structural $\mathcal{L}_S$ and behavioral $\mathcal{L}_B$. We show the resulting training losses in Figure~\ref{fig:train_loss_different_functions}.

We observe that despite our best attempts, both the cross-entropy and distillation losses are numerically unstable for at least one model zoo, whereas the MSE loss remains stable and decreasing in all cases. This could not be resolved by changing the learning rate. For this reason, and because the MSE loss yields already very satisfying results, we decided to focus solely on the MSE loss in this paper.

\subsection{Using the Right Queries Is Essential}
\label{ap:query_set}

While computing the structural loss is a straightforward MSE between the weights of the original model and its reconstruction, comparing their behavior requires using some data, which we have named the \textit{queries} $X = \{ x_i \}_{i=1}^n$. As discussed in Section~\ref{sec:methods}, this raises the question of what data points to use as queries. In this context, we name dataset from which we sample the $n$ queries the \textit{query set}. In this Appendix, we discuss the choice of the query set.

Let a model $f_{\theta_j}$ be trained over some training dataset $(X, y)$. First, since the behavioral loss only compares the outputs of $f_{\theta_j}$ and its reconstruction $f_{\hat{\theta}_j}$, we do not need $y$ and can discard it. We therefore focus on $X$, and its distribution $p(X)$. Indeed, since the input domain $\mathcal{X}$ can be very large, e.g., all pixel values in a $32 \times 32 \times 3$ image, we have guarantees to sample from the subdomain of interest only if we are restricted to a very small subset, e.g., natural images. Similarly, the behavior of $f_{\theta_j}$ is well defined only where $p(X)$ has high probability, as it is the domain it has been trained on. The existence of adversarial examples~\citep{goodfellow2014explaining} shows that even in the close vicinity of $p(X)$, the behavior of $f_{\theta_j}$ is ill-defined. It follows that for our behavioral loss to be effective at reconstructing NNs of similar performance, it needs to be computed as close as possible to $p(X)$.

We experimentally validate this hypothesis. The setup is the same as the one described in Section~\ref{sec:experimental_setup}. We train hyper-representation AEs on all three of our model zoos, using all three of the contrastive $\mathcal{L}_C$, structural $\mathcal{L}_S$ and behavioral $\mathcal{L}_B$ losses. Here, however, we vary the \textit{query set} used when computing the behavioral loss during training. The baseline, which we name ``\textit{Zoo trainset}'', is the same setup as used in Section~\ref{sec:experiments}, where the queries are samples from the training set of the corresponding model zoo. The first variation consists in using natural images, but from a different dataset. This could correspond to a setup where the hyper-representation model does not have access to the zoo's training set and uses some default dataset instead. To that end, we use the STL-10 dataset~\citep{coatesAnalysisSingleLayerNetworks2011}, resized to $32 \times 32 \times 3$, which we name ``\textit{STL-10\_32}''. Finally, we also test using data points sampled uniformly from $\mathcal{X}$, i.e., uniform noise; we name this query set ``\textit{Random}''.

\begin{figure}[t]
    \centering
    \includegraphics[width=\textwidth]{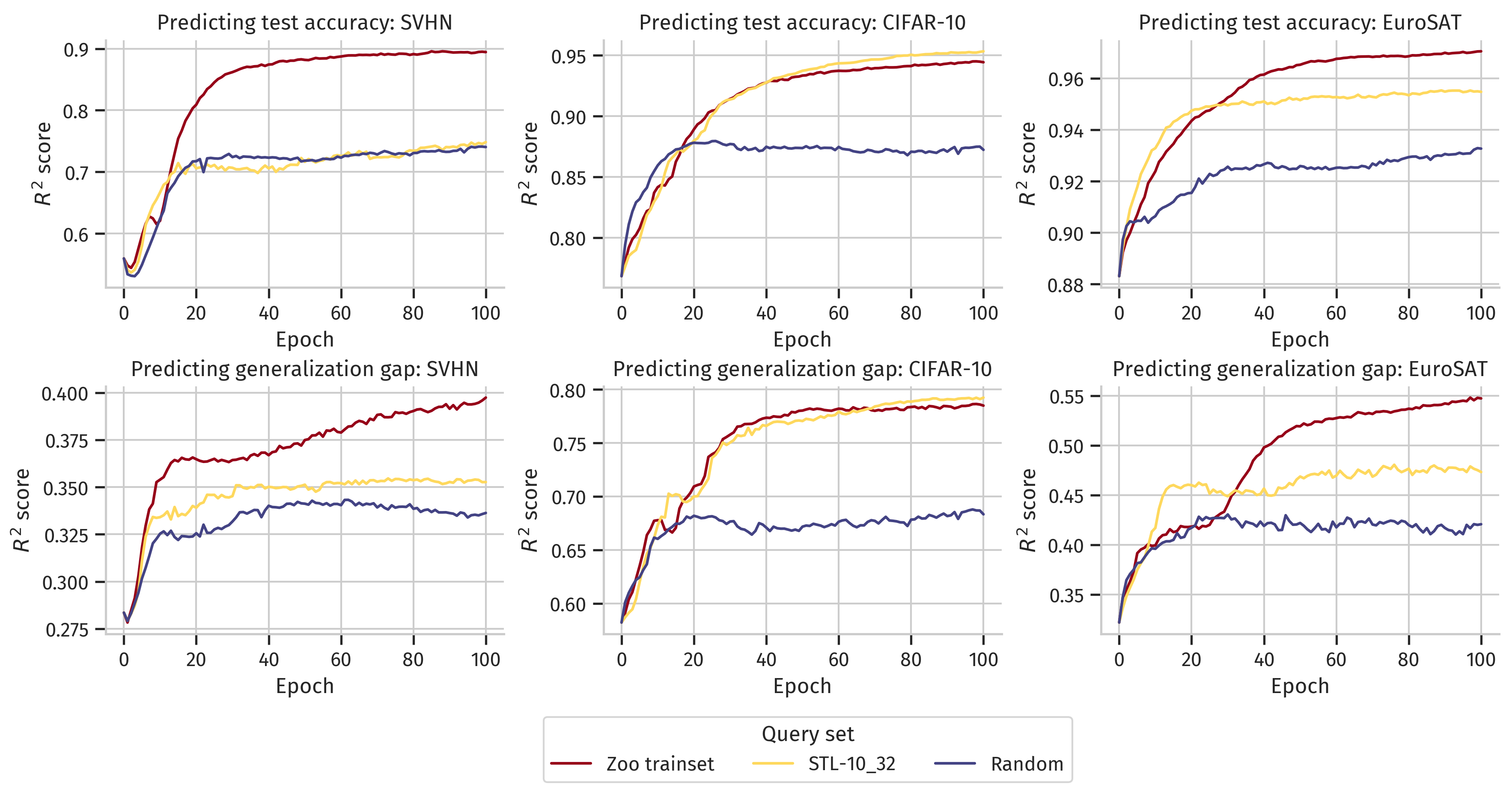}
    \caption{Comparison of the performance of the discriminative downstream tasks by training epoch of \texttt{SANE}, compared for different query sets. \textit{Random} queries yield low performance, and using the \textit{Zoo trainset} yields highest performance for SVHN and EuroSAT. For CIFAR-10, \textit{Zoo trainset} and \textit{STL-10\_32} show a similar level of performance.}
    \label{fig:queryset_discriminative}
\end{figure}

We show the results for the discriminative downstream tasks in Figure~\ref{fig:queryset_discriminative}. As hypothesized, we see large gaps in performance depending on the query set used. In all cases, when using the \textit{Random} query set the performance is low. For both SVHN and EuroSAT, we also see a large gap in performance between \textit{Zoo trainset} and \textit{STL-10\_32}. For CIFAR-10, however, the performance when using \textit{STL-10\_32} is very good, even surpassing \textit{Zoo trainset}. Indeed, STL-10 is a dataset that also contains natural images, and uses similar classes as CIFAR-10. The distributions of SVHN and EuroSAT, respectively house numbers and satellite imaging, are farther in distribution from STL-10.

\begin{figure}[t]
    \centering
    \includegraphics[width=\textwidth]{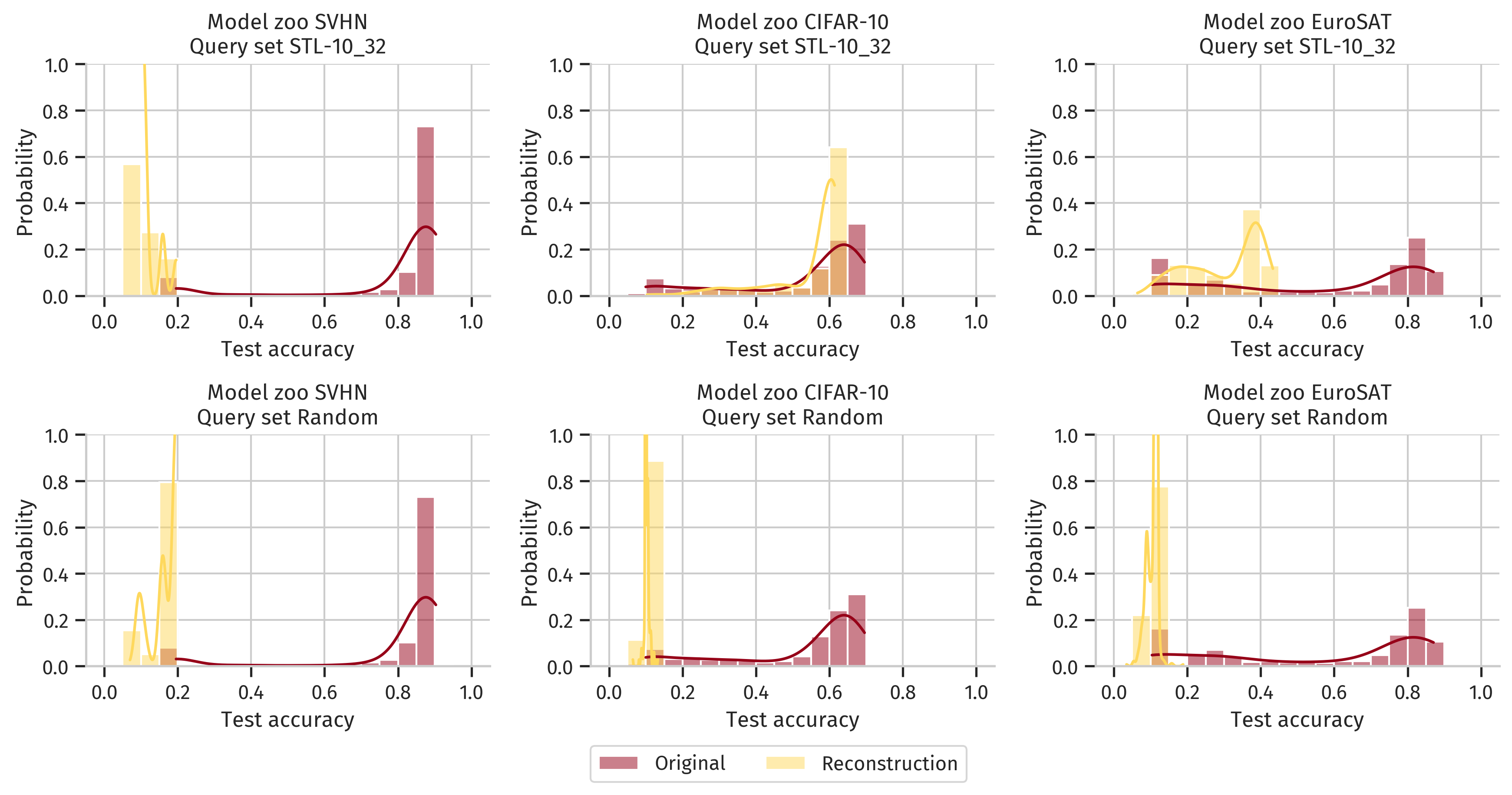}
    \caption{Evaluation of the model reconstructions, shown as distributions of the test accuracy of different models, depending on whether they are part of the original model zoo or reconstructions. We represent variations in the query set in the rows, where \textit{STL-10\_32} and \textit{Random} are represented. Models reconstructed using \textit{Random} queries all perform close to random guessing, and so do SVHN models reconstructed with \textit{STL-10\_32} queries. EuroSAT models reconstructed with \textit{STL-10\_32} queries perform a little better than random guessing but a lot worse than original models. CIFAR-10 models reconstructed with \textit{STL-10\_32} tend to match the distribution of the original models, except the most performant ones, where they fail to perform as well.}
    \label{fig:queryset_reconstructive}
\end{figure}

\newpage

We further show the results for the reconstructive downstream tasks in Figure~\ref{fig:queryset_reconstructive}. When using the \textit{Random} query set, we see a similarly poor performance, with reconstructed models all performing poorly, very close to random guessing. For the SVHN zoo, using \textit{STL-10\_32} performs similarly poorly. It performs slightly better for EuroSAT. When considering CIFAR-10, reconstruction using the \textit{STL-10\_32} query set performs relatively well, but reconstructed models still fail to match the very best models in the zoo in terms of performance.

These results validate our hypothesis that the choice of a right query set is of paramount importance. That is a limitation for the behavioral loss, as contrary to the structural one, it is not data free, and even requires data from the zoo's training set to perform best. Results on CIFAR-10 and STL-10 are, however, encouraging: using a query set that is not the zoo's training set, but that is close enough in distribution, can still yield very satisfying levels of performance. This opens the door for engineering comprehensive query sets for cases where the training set of the models in the zoo is not available.

\subsection{Behavioral Loss Remains Relevant with Limited Computing Resources}
\label{ap:ablation_compute}

As discussed in Appendix~\ref{ap:hr_training}, training hyper-representation models with a behavioral loss takes around twice as much computing time as with only a structural loss. There, we concluded that this was an acceptable trade-off with regard to the increased performance the behavioral loss brings. This raises, however, the question of the validity of the behavioral loss in a compute-constrained setting. In this Section, we evaluate the performance of hyper-representation AEs trained with either the structural or both the structural and behavioral loss, but for a similar computing budget.

To that end, we consider the experiment setup as defined in Section~\ref{sec:experimental_setup}. We compare the hyper-representation model trained with the contrastive $\mathcal{L}_C$ and structural $\mathcal{L}_S$ losses with the one trained with all three, i.e., the one that also includes the behavioral loss $\mathcal{L}_B$. The difference is, we take the hyper-representation $\mathcal{L}_C \oplus \mathcal{L}_S$ after $100$ epochs of training, and compare it to the $\mathcal{L}_C \oplus \mathcal{L}_S \oplus \mathcal{L}_B$ after only $50$ epochs; the computing budget allocated to both is therefore comparable.

When comparing the discriminative downstream tasks' performance in Figure~\ref{fig:discriminative_dstk}, we clearly see that the performance of the hyper-representation trained with $\mathcal{L}_C \oplus \mathcal{L}_S \oplus \mathcal{L}_B$ at $50$ epochs is in all cases higher than that of the one trained on $\mathcal{L}_C \oplus \mathcal{L}_S$ at $100$ epochs. While the performance of the latter seems to increase more rapidly, it reaches a plateau relatively early in the training process; the former, however, keeps improving steadily.

\begin{figure}[ht]
    \centering
    \includegraphics[width=\textwidth]{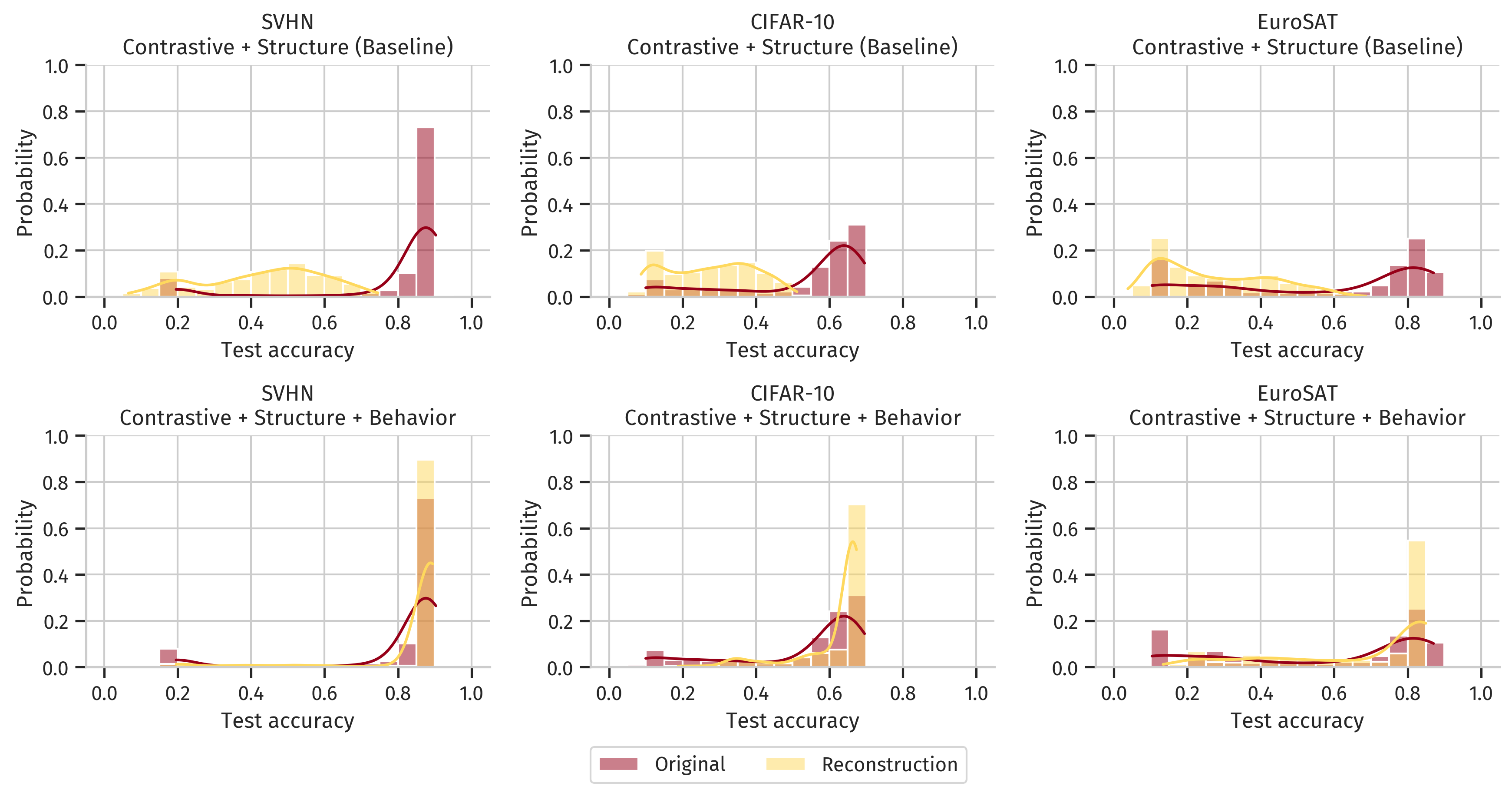}
    \caption{Evaluation reconstructive downstream tasks, for a comparable computing budget where $\mathcal{L}_C \oplus \mathcal{L}_S$ is evaluated after $100$ epochs and $\mathcal{L}_C \oplus \mathcal{L}_S \oplus \mathcal{L}_B$ is evaluated after only $50$ epochs. Even with half the number of training epochs, the AE using the behavioral loss outperforms the one that does not use it. We see, however, that for the EuroSAT zoo, we fail to reconstruct the very best models as performant as they are: the AE could benefit from further training.}
    \label{fig:ablation_compute_reconstruction}
\end{figure}

We further investigate the reconstructive downstream tasks in Figure~\ref{fig:ablation_compute_reconstruction}. There, we see that even with less training epochs, the models reconstructed using the behavioral loss $\mathcal{L}_B$ reach higher levels of performance, comparable to that of the models from the zoo. We note, however, that for the EuroSAT zoo, we fail to reconstruct models as performant as the best models in the zoo: this shows that the AE could still benefit from further training. In general, however, these results indicate that for a fixed computing budget, using both the structural $\mathcal{L}_S$ and behavioral $\mathcal{L}_B$ losses yields better performance than using $\mathcal{L}_S$ only with more training epochs.

\begin{figure}[ht]
    \centering
    \includegraphics[width=\textwidth]{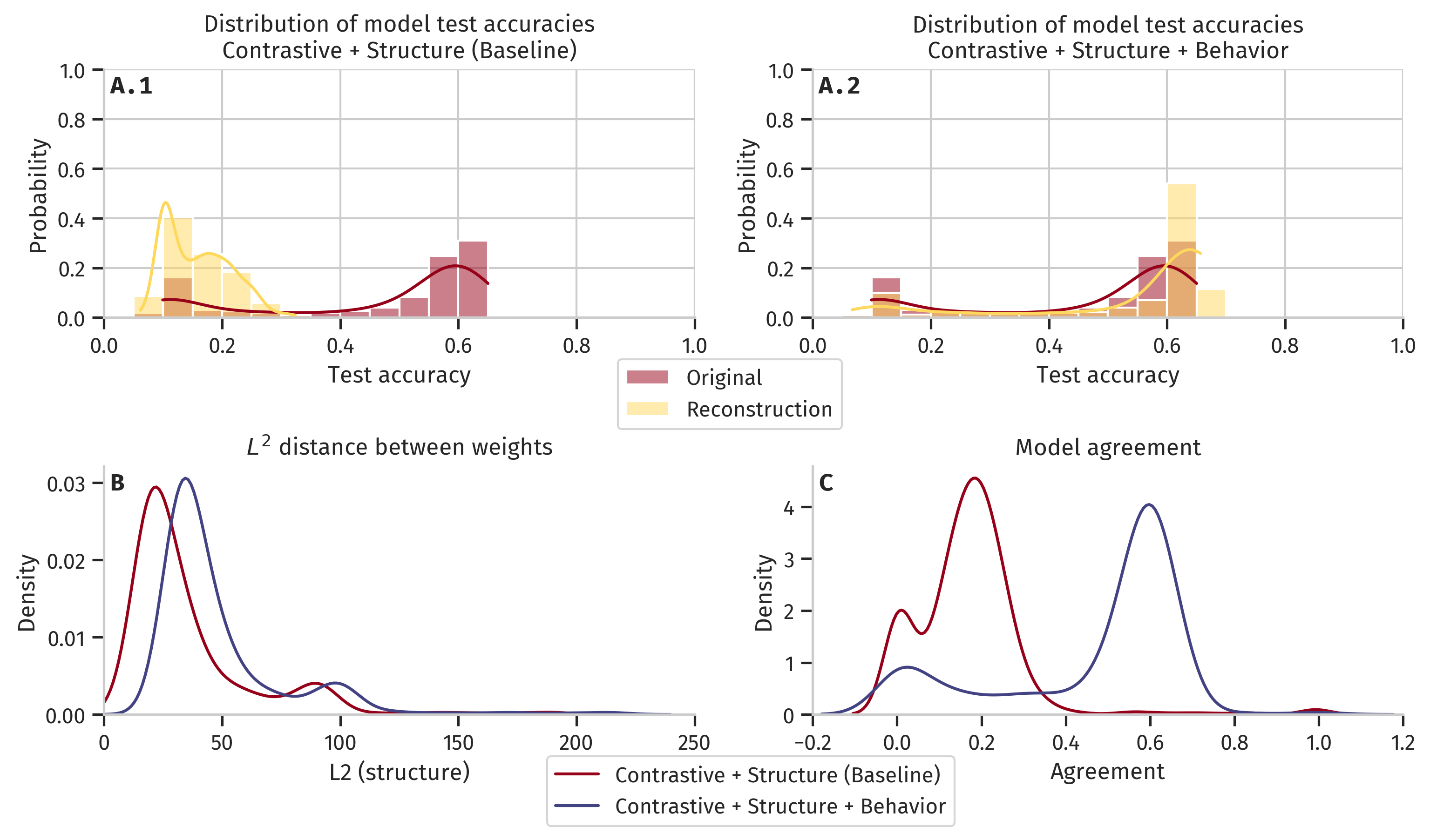}
    \caption{Evaluation of reconstructive downstream tasks, on a larger LeNet-5 architecture, trained on CIFAR-10. Panels A.1 and A.2 show the distribution of test accuracies of models and their reconstruction. They show that the distribution of reconstructed models' performance is much closer to that of the originals' when using both $\mathcal{L}_S$ and $\mathcal{L}_B$. Panel B shows the pairwise $L^2$ distances between the weights of models and their respective reconstruction; we note a similar distribution with slightly lower distances when using $\mathcal{L}_S$ only. Panel C shows the pairwise model agreement between models and their respective reconstruction, demonstrating a higher level of agreement when using both $\mathcal{L}_S$ and $\mathcal{L}_B$.}
    \label{fig:cifar10_lenet5}
\end{figure}

\subsection{Combining Structure and Behavior Works on Larger Architectures}

In this Appendix, we explore whether our findings generalize to a larger CNN architecture. To do so, we build a new model zoo using the same hyperparameters as those described in Appendix~\ref{ap:model_zoos_generation} but based on the LeNet-5~\citep{lecunGradientBasedLearningApplied1998} architecture, for the CIFAR-10 dataset. This architecture has $62,006$ parameters, or roughly $6$ times as many parameters compared to our other zoos. We show results in Figure~\ref{fig:cifar10_lenet5}, which validate that using $\mathcal{L}_C \oplus \mathcal{L}_S \oplus \mathcal{L}_B$ outperforms the baseline ($\mathcal{L}_C \oplus \mathcal{L}_S$) for reconstructive downstream tasks.

\newpage

\subsection{Selecting the Weight of the Structural Loss}
\label{ap:selecting_beta}

In this Appendix, we describe how we selected the hyperparameter $\beta$, which is used to control the weight of the structural loss $\mathcal{L}_S$ and the behavioral loss $\mathcal{L}_B$. To do so, we used the validation set of our model zoo, and evaluated the discriminative and reconstructive downstream tasks on it using different values of hyperparameter $\beta$.

We show the results for the discriminative downstream tasks in Table~\ref{tab:discriminative_dstk_val}. There, a value of $\beta = 0.1$ shows best performance across all datasets, except when predicting the generalization gap on CIFAR-10. Differences between the performance of the downstream tasks are larger when considering the SVHN dataset than for the other two. For these reasons, with regard to the discriminative downstream tasks, we select a value of $\beta = 0.1$.

We show the results for the reconstructive downstream tasks in Table~\ref{tab:reconstructive_dstk_val}. As can be expected, a larger $\beta$, which is linked with a larger weight for the structural loss $\mathcal{L}_S$, generally leads to lower $L^2$ distance between model weights. When considering model agreement, however, a value of $\beta = 0.1$ seems to perform best across all datasets. Since this value performs best for both discriminative and reconstructive downstream tasks, we select it for the rest of our experiments.

\begin{table}[h]
    \caption{Discriminative downstream tasks performance on the validation set, for various values of hyperparameter $\beta$. We predict the test accuracy and generalization gap of our models based on their latent representation, using a linear probe. We give the $R^2$ score for predictions on the held-out test split. For the SVHN dataset, using $\beta = 0.1$ clearly outperforms all other options. When predicting model test accuracy, it is also the best option, by a small margin. Results for predicting the generalization gap are more inconclusive for CIFAR-10 and EuroSAT, but differences in performance remain relatively small.}
    \label{tab:discriminative_dstk_val}
    \begin{center}
    \tabcolsep=0.10cm
    \begin{tabular}{lcccccc}
    & \multicolumn{3}{c}{\textsc{\textbf{Test accuracy}}} & \multicolumn{3}{c}{\textbf{\textsc{Generalization gap}}}\\
    \multicolumn{1}{c}{\textsc{\textbf{$\beta$}}} & \textsc{SVHN} & \textsc{CIFAR-10} & \textsc{EuroSAT} & \textsc{SVHN} & \textsc{CIFAR-10} & \textsc{EuroSAT}\\
    \midrule 

    0.0 & 0.725 & 0.932 & 0.940 & 0.338 & 0.799 & \textbf{0.622} \\
    0.1 & \textbf{0.890} & \textbf{0.964} & \textbf{0.963} & \textbf{0.429} & 0.788 & \textbf{0.622} \\
    0.2 & 0.667 & 0.958 & 0.957 & 0.295 & 0.789 & 0.576 \\
    0.3 & 0.657 & 0.959 & 0.962 & 0.290 & 0.776 & 0.557 \\
    0.4 & 0.668 & 0.958 & 0.948 & 0.324 & 0.806 & 0.578 \\
    0.5 & 0.701 & 0.941 & 0.943 & 0.337 & \textbf{0.807} & 0.536 \\
    
    \end{tabular}
    \end{center}
\end{table}

\begin{table}[h]
    \caption{Reconstructive downstream tasks performance on the validation set, for various values of hyperparameter $\beta$. We evaluate structural reconstruction with the average $L^2$ distances between the weights of test split models and their reconstructions. We evaluate behavioral reconstruction with the average classification agreement between test split models and their reconstructions. Standard deviation is indicated between parentheses. When considering structural distance, a larger $\beta$ seems linked with a lower distance. When considering model agreement, a value of $\beta = 0.1$ performs best across all three datasets.}
    \label{tab:reconstructive_dstk_val}
    \begin{center}
    \tabcolsep=0.10cm
    \begin{tabular}{lcccccc}
    & \multicolumn{3}{c}{\textsc{\textbf{Structure ($L^2$ distance)}}} & \multicolumn{3}{c}{\textbf{\textsc{Behavior (Agreement)}}}\\
    \multicolumn{1}{c}{\textsc{\textbf{$\beta$}}} & \textsc{SVHN} & \textsc{CIFAR-10} & \textsc{EuroSAT} & \textsc{SVHN} & \textsc{CIFAR-10} & \textsc{EuroSAT}\\
    \midrule 

    0.0 & 56.0 (±6) &   61.0 (±9) &  64.2 (±16) & 19.6\% (±1\%) &   10.5\% (±4\%) &   10.8\% (±4\%) \\
    0.1 & 27.1 (±9) &  30.0 (±12) &  40.6 (±20) & \textbf{82.1\% (±18\%)} &  \textbf{54.5\% (±13\%)} &  \textbf{66.6\% (±23\%)} \\
    0.2 & 26.7 (±8) &  28.3 (±12) &  39.5 (±20) & 64.9\% (±18\%) &  53.5\% (±16\%) &  64.9\% (±25\%) \\
    0.3 & 26.3 (±8) &  27.9 (±11) &  37.4 (±18) & 65.2\% (±18\%) &  51.4\% (±18\%) &  63.9\% (±26\%) \\
    0.4 & \textbf{25.5 (±8)} &  \textbf{27.5 (±10)} &  37.0 (±18) & 65.0\% (±18\%) &  50.6\% (±18\%) &  62.8\% (±26\%) \\
    0.5 & 25.6 (±8) &  28.0 (±10) &  \textbf{35.9 (±18)} & 64.9\% (±18\%) &  47.5\% (±16\%) &  61.4\% (±26\%) \\
    
    \end{tabular}
    \end{center}
\end{table}

\end{document}